**RESEARCH ARTICLE**                                                                                   **OPEN ACCESS**

# An Efficient Method of Partitioning High Volumes of Multidimensional Data for Parallel Clustering Algorithms

Saraswati Mishra[1], Avnish Chandra Suman[2]
[1]Centre for Development of Telematics, New Delhi - 110030, India. saraswatimishra18@gmail.com
[2]Centre for Development of Telematics, New Delhi - 110030, India. avnishchandrasuman@gmail.com

**ABSTRACT**
An optimal data partitioning in parallel/distributed implementation of clustering algorithms is a necessary computation as it ensures independent task completion, fair distribution, less number of affected points and better & faster merging. Though partitioning using Kd-Tree is being conventionally used in academia, it suffers from performance drenches and bias (non equal distribution) as dimensionality of data increases and hence is not suitable for practical use in industry where dimensionality can be of order of 100's to 1000's. To address these issues we propose two new partitioning techniques using existing mathematical models & study their feasibility, performance (bias and partitioning speed) & possible variants in choosing initial seeds. First method uses an n-dimensional hashed grid based approach which is based on mapping the points in space to a set of cubes which hashes the points. Second method uses a tree of voronoi planes where each plane corresponds to a partition. We found that grid based approach was computationally impractical, while using a tree of voronoi planes (using scalable K-Means++ initial seeds) drastically outperformed the Kd-tree tree method as dimensionality increased.
***Keywords:*** Clustering, Data Partitioning, Parallel Processing, KD Tree, KD-Tree, Voronoi Diagrams

## I. INTRODUCTION

While profiling a hybrid (parallel & distributed) implementation of OPTICS (Goel et all., 2015) algorithm we had an observation that over 50% our threads were outperforming the rest by huge margins. Our method was to 1. partition existing data into x parts 2. treat each part as a separate input and run the parallel OPTICS thread on each 3. merge the resultant clusters. We used Kd – tree (Bentley, 1975) to make partitions. K-d tree is a variance of binary tree where each node represents a data point in n-dimensional space. Every leaf node of k-d tree represents a splitting of a (n-1) dimensional hyper-plane resulting in two half-planes which can be thought of as partitions. Following method was used.

1. Start with dimension having highest variance and divide data set based on value points in that dimension only. Result is two different partitions.
2. Repeat the process with next dimension of highest variance until desired numbers of partitions are made.

A sample representation will look something like this

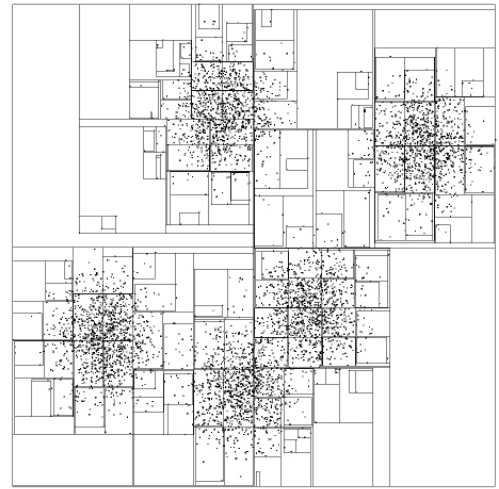

**Figure 1 :** Kd-Tree Partitioning

We observed various limitations. Assume x partitions, n data points, m partitions, d dimensions
1. Data scan was needed in every stage i.e. for getting m partitions we needed O(m) scans over same points again and again.
2. Median finding was a costly operation of order O(n). However when executed for every non-leaf node , the overall cost was of order O(mn)
3. K-d tree doesn't guarantee considering every dimension. In-fact there it exhibits a bias towards a few dimensions of high variance, hence points in a partition can be relatively far than points in different dimensions for a large d. This will lead to a poor clustering result. Also as the number of dimensions tend to increase, the performance of k-d tree with regards to bias decreases.
4. Partitions are half planes. They have a general tendency to result in more number of affected points in uniformly or near-uniformly distributed space.
5. No inherent merging structure. A merging strategy needs to be implemented

The most concerning of these at that time for us was 1 & 2 as they were highly serial and slow part of our parallel implementation. We wanted to optimize the multiple passes over data to as few as possible. Also we wanted a better way of finding median or near median. In next section we will see how we tried solving 1 & 2 by splitting our d-dimensional space into d-cubes such that in a single pass we determine the points in every cube & their cardinality and other computations which improves median finding performance significantly. However we soon realized the impractical aspects of our approach and it inherently lacking the solutions to 3, 4 & 5. We moved on to next approach where we partition our space using n-dimensional voronoi diagrams and found it to be extraordinarily outperforming kd-tree with proper initial seeds.

## II. N-CUBE BASED APPROACH

The proposed partitioning approach works by initially dividing the n-dimensional space into n-Cubes by making y splits along each dimension so that we obtain $(y+1)^n$ cubes.

The basic idea here is to create an overall index (a summary in space based on cubes) such that for any partitioning computation, we only need to use this summary and not the entire data. n-Cubes work as following

*Algorithm: Construct Cubes*
*Input: Set of x points p in space $(p_1,p_2...p_x)$ where each $p_x$ is $(x_1,x_2..x_n)$.*
*Number of partitions m*
*k: a natural number $1<=k$*
*#higher values for k ensures better overall distribution but lower performance.*

*Output: Set of Cubes $(c_1, c_2 .... c_m)$*
    *Where $m = (y+1)^n$*
*Procedure:*
*# Finding an optimal y & initializing cube boundaries*
*Let $minx_n$ be min $(p_x)$ in $n^{th}$ dimension*
*Let $maxx_n$ be max $(p_x)$ in $n^{th}$ dimension*
*Let $Y = ky$. Let $M = (Y+1)^n$*
*for each $c_i$ in $(c_1..c_M)$*

*Boundary$(c_{Mn})$ =*
    *$\{(i-1)*(max_n-min_n)/M , i*(max_n-min_n)/M\}$*
*Binary sort c*
*For each pi in $(p_1...p_x)$*
*Find pi's location in p.*
*Add $p_i$ to $c_M$*
*$c_M$.totalPonits++*

*Algorithm: Find Median*
*Input : Set of Cubes $(c_1...c_M)$, Total Points*
*Output : Median along a dimension n say $m_n$*
*Procedure :*
*P : total points in space*
*x=0;*
*While x<P/2*
*Move to next cube , $x = x+C_m$.totalPoints*
*#We stop at the cube that contains our median (or a close approximation if k is too low or too high)*
    *Find median of set of points in $C_m$.*

We can see that cube creation is an $O(n+logn)$ task while median creation is an $O(n/M)$ task which looks very efficient. The approach should have worked in two passes over data plus a single pass on grid cells. However there were major design & implementation issues with this approach
1. The number of cubes is $(y+1)^n$. Even if y is 2, for a huge n, we get $2^n$ cells. This grows closer to total number of data points.
2. Programmatically unfeasible in direct sense. Accessing n-dimensional arrays needs n loops, we don't know n beforehand. Solution is to convert n-d cells to 1-d (resulting in very long 1-d array). Unable to allocate memory on stack for the 1-d array.
3. Too many cells, sparse cells, data distribution across cells not uniform at all.
4. Most of the partitions will be empty, even when the number of data points N is large, leading to extreme waste of memory and CPU time.
5. Hence we conclude that the method was not Suitable for > 2d data
6. Didn't solve problem of bias but tends to worsen it with a higher cubes number

## III. VORONOI DIAGRAM BASED PARTITIONING

Our second approach involves voronoi diagrams (Aurenhammer,1991). Our goal is to construct a partitioning scheme that handles high-dimensionality as well and not just provide good performance by ignoring a lot of dimensions. It is also necessary that partitions should not hold too many or too few points. Number of passes on data should be as less as possible preferably ~1.

An efficient way to satisfy problem 1 & 2 of kd-tree is a tree structure that needs $O(\log n)$ number of comparisons on average to distribute a point and to determine the affected partitions where n is the number of nodes in the tree. One way to satisfy problem of considering all dimensions all together is to use a Voronoi diagram which partitions the space into Voronoi cells, directed by a set of split points $Q = q_1, q_2, \ldots$ such that for each cell corresponding to split point $q_i$, the points x in that cell are nearer to $q_i$ than to any other split point in Q. Hence, by constructing a tree of Voronoi diagrams, we can satisfy our two major concerns.

Let's call such a structure as v_tree. The top or root node of a v_tree gives a brief summary of the whole data and is split to many Voronoi cells which are split as well and so on.

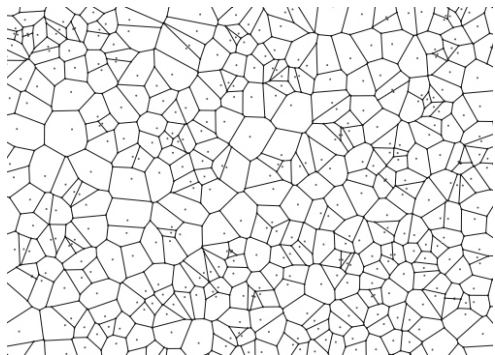

**Figure 2:** A voronoi diagram depicted in two dimensions

**Construction of v_tree**
At each level, find k (2) centre
1. Points must be appropriately spaced and far
2. Use scalable kmeans++ (Bahmani et all., 2012) or gnat technique (Brin ,1995)
3. Assign all points to either centre based on distance
4. All points that lie in current to current + eps boundary (Goel et all., 2015) are considered affected.
5. Delete extra points stored in parent node to remove redundancy

Repeat for new sets until requisite number of partitions found.

**Data Structure**
A minimal n-voronoi-tree implementation data structure in c will look like this.
```
typedef struct member {
int id;
float *val;
} member;
struct v_node {
int level;
int core1,core2;
member *mem1,*mem2,*mem_overlap;
struct v_node *left,*right;
int     count_mem1,count_mem2,    count_overlap,
total_count;
}
struct v_tree {
int levels;
struct node *head;
}
```

Head node is the summary of Entire Data. Each parent node contains summary of points in child nodes. The exact points are stored until data is partitioned at level & deleted as soon as we move to next level.

Note that v_tree might not necessarily be binary. More than two centres can be chosen.

The leaf nodes are our final partitions, and going up the tree inherently makes a merging structure for resultant clusters. Load in each partition will depend upon the center chosen and distribution of data in n-dimensional space.

**How to choose initial seeds?**
1. Select randomly: Choosing centre randomly doesn't guarantee or breach anything and simply leaves thing to the centre chosen and distribution of data. There is equal probability of getting each load distribution. Hence the probability of getting a perfect load balance tends to zero.
2. GNAT Approach: Suppose we need n seeds. We start by selecting a point at random. The next point is chosen such that its distance from first point is maximum, the third point is chosen such that its distance from sum of previous two points is maximum. Similarly forth point should be the point farthest from sum of first three points and so on. This approach is good in terms of load balancing for a big value of n, but cannot guarantee good balance for small n (~2-5).
3. Scalable K-Means++ Approach: Suppose k centre are needed, C be the set of initial seeds, then
a. Sample a point uniformly at random from the data points.
b. For each data point p, compute it's distance from nearest centre.
c. Choose a m point p using Weighted probability distribution where a point p is chosen with probability proportional to $D(p)2$. Assume it be new centre
d. If you have k centres, proceed with partitioning, else repeat 2 & 3

The centres that we get tend to be close to the centroids of clusters present in data, assuming there are k-clusters. Load distribution is entirely dependent on the data; however we can be optimistic about not getting very biased distribution with real-life datasets.

Such a partitioning hugely tends to reduce the time spent in merging the final clusters as there will be minimal affected points.
4. More than 2 centre: can be used in implementation to satisfy various criteria.
a. All the above approaches (or any random or probability based approach) will tend to good balancing as we increase number of centre. This can be proved mathematically using induction. We receive a perfect balance when numbers of points equals number of centre i.e. each point is a partition.
b. In a case when number of partitions is not in power of 2.
c. Different number of centre at different level can be used for perfect guided partitioning.
5. How about Median: Points very close to median and on same axis as centre will produce exact partitions (similar to kd-tree), however complexity sumps up to O (kd-tree) +O (v_tree).

## IV. EXPERIMENTATION & RESULTS

We used following environment to execute test & profile. Execution Environment: - Ubuntu 13.04, Intel Core i3 (3rd generation), 2.4GHz (2 cores hyper threaded), 4GB RAM, 3MB L2 Cache. Compilation Environment: - C, gcc, gdb, gprof, vampir, Geany IDE .Visualization of output was done using geogebra.

Test Data-sets: (no. of points x no. of dimensions, double precision data)
1. 100x2,
2. 700x9,
3. 1500x1024,
4. 4000x1024,
5. 40000x1024

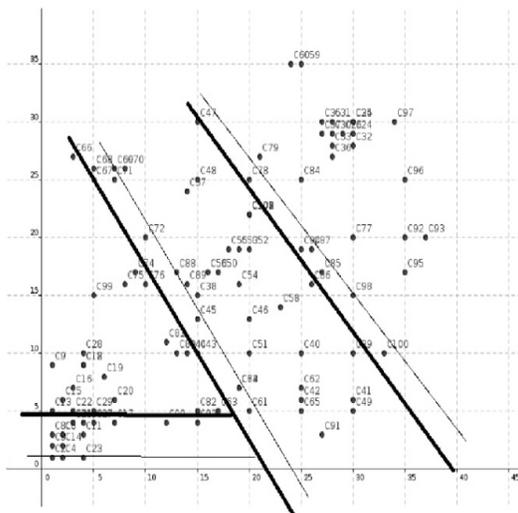

**Figure 3 :** Comparison of v_tree partitioning result with four partitions

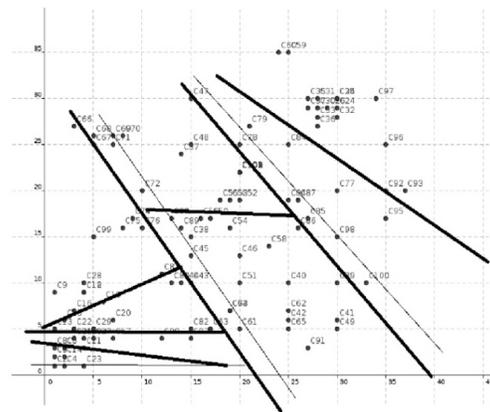

**Figure 4:** Comparison of v_tree partitioning result with eight partitions

We noticed an increase in load bias as number of partitions increase.

Distribution was almost uniform with uniform data. However we can see that Kd-tree (right) provides a better distribution with uniformly distributed data. However this will decrease with number of dimensions.

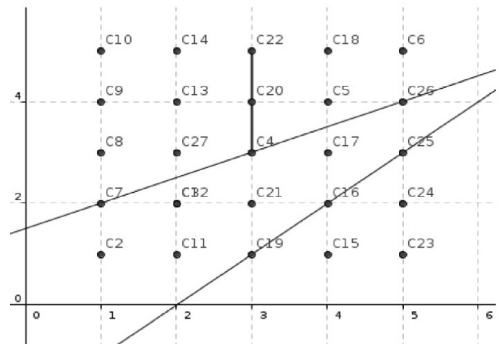

**Figure 5 :** Comparison of uniform data partitioning (v_tree )

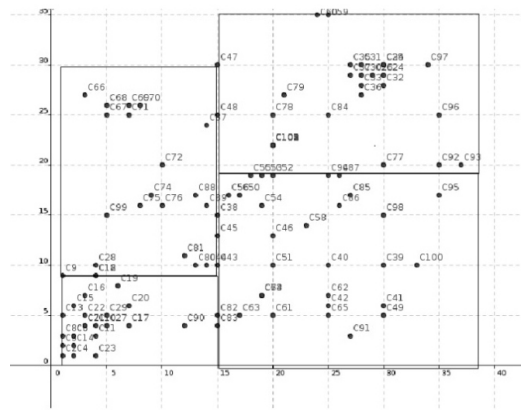

**Figure 6** Comparison of uniform data partitioning (k-d tree)

**Performance** (in approx ~10sec units)

| Data Set  Approach | 700*9 | 1500*1024 | 4000*1024 | 40000*1024 |
|---|---|---|---|---|
| Kd - Tree | 0.02 | 9.88 | 46.67 | 512 |
| v_tree (Scalable Kmeans++) | 0.01 | 3.05 | 7.61 | 73.34 |
| v_tree (Median) | 0.02 | 9.96 | 54.43 | 542 |

## V. CONCLUSION

We have seen that for higher dimensionality the approach proposed takes very less time as compared to kd-tree approach; however v_tree technique needs a load balanced variant to boast perfect results.

Our future works will include better load balancing, comparison of merging time and distribution time, parallelizing the approach and implementing a grid based alternative.

## APPENDIX

Related Code, Data Sets & Results can be requested from
https://drive.google.com/file/d/0Bxo9wQ432jhla1dQNWFrbnM4bnc/view?usp=sharing